\pdfoutput=1

\documentclass[11pt]{article}

\usepackage[final]{ACL2023}

\usepackage{times}
\usepackage{latexsym}

\usepackage[T1]{fontenc}

\usepackage[utf8]{inputenc}

\usepackage{microtype}

\usepackage{microtype}
\usepackage{color}
\usepackage{times}
\usepackage{latexsym}
\usepackage{url}
\usepackage{booktabs}
\usepackage{tikz}
\usepackage{anyfontsize}
\usepackage{caption}
\usepackage{subcaption}
\usepackage{xstring}
\usepackage{todonotes}
\usepackage{csquotes}
\usepackage{pifont}
\usepackage{mathtools}
\usepackage{xcolor}
\usepackage{enumitem}
\usepackage{diagbox}
\usepackage{inconsolata}
\usepackage{outlines}

\newcommand{\wtan}[1]{{\color{blue}\bf\small [wtan: {#1}]}}

\definecolor{lightblue}{HTML}{3cc7ea}

\definecolor{lightgreen}{HTML}{90EE90}

\newcommand{\spa}{SPA}
\newcommand{\spanew}{{\sc PostText}}

\title{Reimagining Retrieval Augmented Language Models\\for Answering  Queries\\{\small [Reality Check Theme Track]}}

\author{Wang-Chiew Tan\ \ Yuliang Li\ \ Pedro Rodriguez\\ {\bf Richard James\textsuperscript{*}\ \ Xi Victoria Lin\ \
Alon Halevy\ \ Scott Yih} \\
  Meta \\
  \texttt{\{wangchiew,yuliangli,victorialin,ayh,scottyih\}@meta.com\ \ rich@richjames.co}\textsuperscript{*}}

\begin{document}

\maketitle

\begin{abstract}
We present a reality check on large language models and inspect the promise of retrieval-augmented language models in comparison. Such language models are semi-parametric, where models integrate model parameters and knowledge from external data sources to make their predictions, as opposed to the parametric nature of vanilla large language models. We give initial experimental findings that semi-parametric architectures can be enhanced with views, a query analyzer/planner, and provenance to make a significantly more powerful system for question answering in terms of accuracy and efficiency, and potentially for other NLP tasks. 
\end{abstract}

\section{Introduction}
As language models have grown larger~\citep{kaplan-etal-2020,hoffmann2022scale}, they have fared better and better on question answering tasks~\citep{hendrycks2021measuring} and have become the foundation of impressive demos like ChatGPT~\citep{ouyang2022training,chatgpt3}.
Models like GPT-3~\citep{brown2020gpt3} and ChatGPT generate fluent, human-like text, which comes the potential for misuse as in high-stakes healthcare settings~\citep{dinan2021safety}.
Large language models (LLMs) also come with several significant issues~\citep{hoffmann2022scale, bender-etal-2021}.  


LLMs are costly to train, deploy, and maintain, both financially and in terms of environmental impact~\citep{bender-etal-2021}. These models are also almost always the exclusive game of industrial companies with large budgets.
Perhaps most importantly, the ability of LLMs to make predictions is not commensurate with 
their ability to obtain insights about their predictions.
Such models can be prompted to generate false statements~\citep{wallace2019universal}, often do so unprompted~\citep{asai2022guided} and when combined with its ability to easily fool humans, can lead to misuse~\citep{gpt3-reddit}.


In recent years, we have seen the promise of retrieval-augmented language models 
partially addressing the aforementioned shortcomings~\citep{guu2020realm,lewis2020rag,borgeaud2021retro,izacard2022atlas,michihiro22:racm3}. The architecture of such models is {\em semi-parametric}, where the model integrates model parameters and knowledge from external data sources to make its predictions. 
The first step of performing a task in these architectures
is to retrieve  relevant knowledge from the external sources, and then perform finer-grained reasoning. Some of the benefits these architectures offer are that the external sources can be verified and updated easily, thereby reducing hallucinations~\citep{shuster-etal-2021} and making it easy to incorporate new knowledge and correct existing knowledge without needing to retrain the entire model~\cite{lewis2020rag}. 
Models that follow semi-parametric architectures (\spa) are typically smaller than LLMs and they have been shown to outperform LLMs on several NLP tasks such as open domain question answering (see Table~\ref{tbl:comparison}). 
Recent work that extends LLMs with modular reasoning and knowledge retrieval~\citep{mkrl-2022,langchain} is also a type of SPA. 
\begin{table*}[th]
\centering
\small
\begin{tabular}{cccc} \toprule
Model & \#Params & Outperformed LLM's sizes         & Tasks              \\ \midrule
REALM~\citep{guu2020realm} & 330M     & 11B (T5)                         & Open-QA                 \\
RETRO~\citep{borgeaud2021retro} & 7.5B     & 178B (Jurassic-1), 280B (Gopher) & Language modeling       \\
Atlas~\citep{izacard2022atlas} & 11B      & 175B (GPT-3), 540B (PaLM)        & Multi-task NLU, Open-QA \\
RAG~\citep{lewis2020rag}   & 400M     & 11B (T5)                         & Open-QA                 \\
FiD~\citep{izacard2021fid}   & 770M     & 11B (T5), 175B (GPT-3)           & Open-QA                \\ \bottomrule
\end{tabular}
\caption{The sizes of SPA models with those of comparable or outperformed LLMs.}
\label{tbl:comparison}
\end{table*}


In this paper we argue that
building on the core ideas of \spa{}, we can potentially construct much more powerful question answering systems that also provide access to multi-modal data such as image, video and tabular data. 
We describe \spanew{}, a class of systems that extend \spa{} in three important ways. First, \spanew{} allows the external data to include {\em views}, a concept we borrow from database systems~\citep{dbcompletebook}.
A {\em view} is a function over a number of data sources, $V$ = $f(D_1,...,D_n)$. In databases, SQL queries are used to define tabular views. For example, $V$ can be a table of records of minors that is derived from a table of person records by selecting only those with age$<$18.
In general, however, views need not be tabular. When a view is materialized (i.e., executed and stored), it may be useful for answering certain queries\footnote{We use queries and questions interchangeably.} more effectively. In this paper, we adopt a more general notion of views, not limited to results of SQL queries, which can (compositionally) support a variety of user questions.
Views are particularly important to support multi-modal data, because combinations of data from multiple modalities can be modeled as views. 
Second, \spanew{} contains
a question analyzer and planner module that decides on the best strategy to answer a question that may involve first answering multiple subquestions in sequence or in parallel. 
This module bears similarity to query optimization techniques in database systems but will go significantly beyond the techniques established in database systems 
since, 
there are multiple different ways to answer a natural language question, especially with the availability of multi-modal data. 
Finally, \spanew{} supports computing the provenance of answers to questions. 
The provenance-aware answer generator module can track the evidence (training data or external sources) that is used for the answers, even if views are used as intermediate results. 

We illustrate   
the power of \spanew\ with examples in the next section and also the overview of its architecture. In the remaining sections, we describe the different components of \spanew.
{\bf add some description of experiments}

\section{Overview of PostText}
\label{sec:examples}


\begin{figure*}
\centering
\includegraphics[height=3.2in, width=0.85\textwidth]{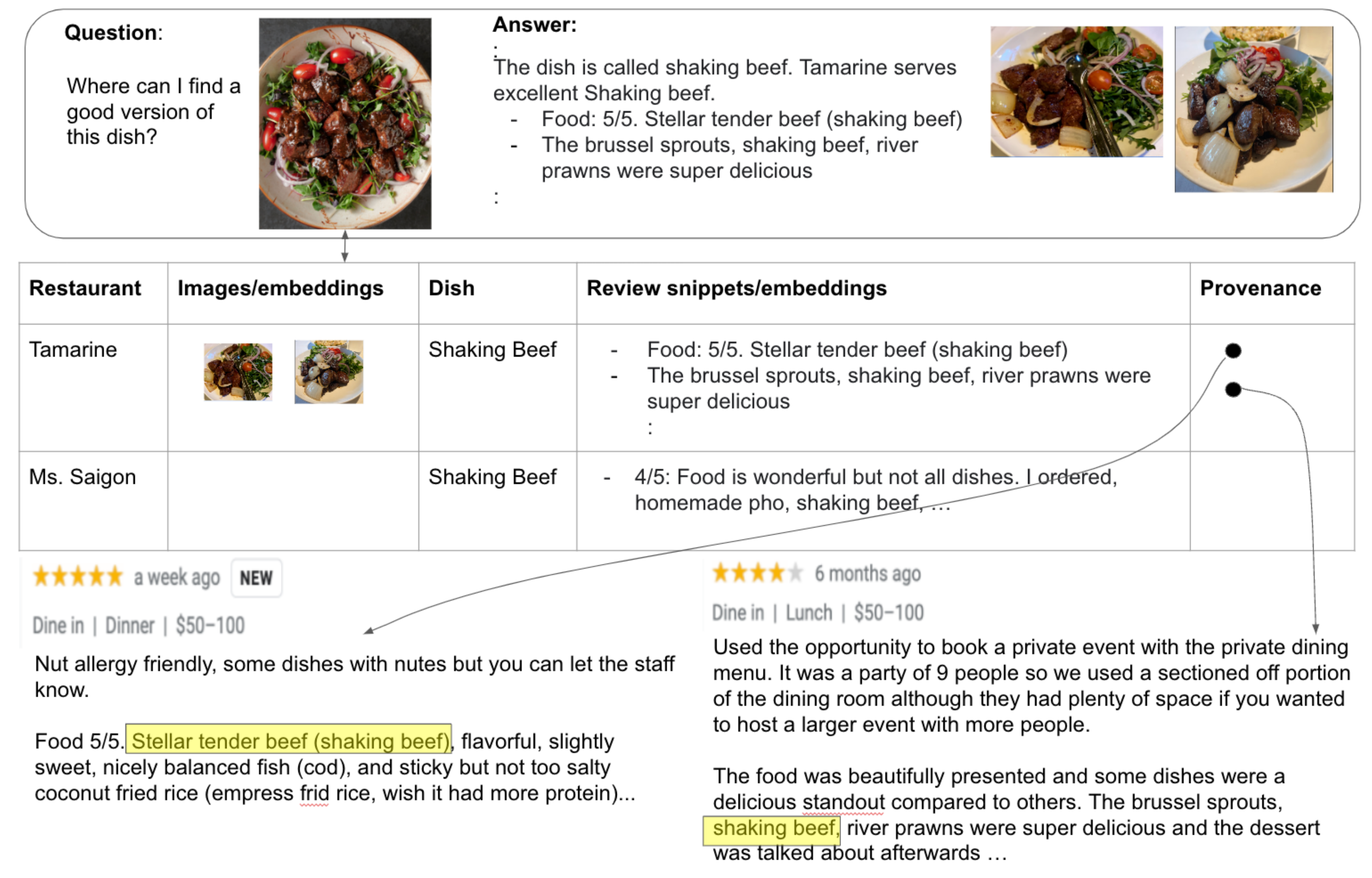}
\caption{Multimodal question with multimodal answer. The view (middle) associates the dishes with its corresponding review snippets and images. The provenance links show where the snippets are extracted from. There are also provenance links for the images and name of the dish (not shown).}
\label{fig:example1-multimodalquestion}
\end{figure*}


\noindent
{\bf Example 1~} Consider a setting where we answer questions over data that includes images of dishes and text with restaurant reviews. We can create a view that aligns these two data sets so we can answer more complex queries readily. The view, the table in the middle of Figure~\ref{fig:example1-multimodalquestion}, aligns dishes with relevant reviews and the corresponding restaurants. Note that creating this view involves  an intermediate step of identifying the name of the dish in an image. The view also stores the 
provenance links to the actual reviews from which the snippets were extracted. There are also provenance links for the images and the name of the dish (not shown).

This view can be used to answer questions that would be more difficult without it. For example, if a person recalls a nice dish she had in the past but does not remember its name and is trying to figure out which restaurants serve the same dish and what are the reviews, she can pose the question, which includes both the question in text and an image of the dish. The answer states the name of the dish in question and lists restaurants with top reviews for that dish, along with images of the dish and snippets of those reviews and their provenance.

\smallskip
\noindent
{\bf Example 2~} The same 
view can also be used to answer the question ``{\em how many reviews raved about 
Shaking beef?}''. The answer requires counting the number of reviews that are synonymous to very positive reviews about Shaking beef.
The view surfaces the reviews associated with Shaking beef immediately and alleviates the amount of work that is required to compute the answer otherwise.

\smallskip
The examples show that some questions can be answered more easily if they 
are supported by views that surface useful associations between data. 
In fact, indices are a type of views to accelerate lookups between an item and its attributes. In database systems, views have been used extensively to enable more efficient query answering~\citep{halevy-views-2001,GL-2001} with significant work on automatically materializing a set of indices for efficient query answering~\citep{jindal-etal-2018selectviews,das-etal-2019msazure}.
A set of views and indices are defined automatically or manually in anticipation of a set of frequently asked 
queries under a budget constraint, e.g., space, so that during runtime, most of the incoming queries can be answered immediately or after applying simple operations over the views. Otherwise, the system falls back to answering the queries using the actual data sources.
In other words, \spanew\ prefers to use views to answer the questions, which will likely to be more efficient and accurate in general but otherwise, the system falls back to the traditional question answering strategy.
In addition to query answering, views have also been used to define content-based access control~\citep{bertinosandhu-2005}, i.e., which parts of the data are accessible and by whom.

The examples also show how provenance is provided as part of the answer.
In these examples, it happened that provenance was easily determined through the provenance links that are already captured in the views.
If actual data sources are accessed, the links to the data sources used (e.g., spans of text documents, parts of images, segments of videos) to derive the answer are provided as part of the answer.
If the answer is generated by the language model, we trace how \spanew{} derives the answer from parametric knowledge and retrieved data through analyzing its weights or determining ``influential'' parametric knowledge (Section~\ref{sec:gen}) similarly to~\cite{akyurek2022tracing}.


\begin{figure*}[!t]
    \centering
    \includegraphics[height=2.4in, width=0.83\textwidth]{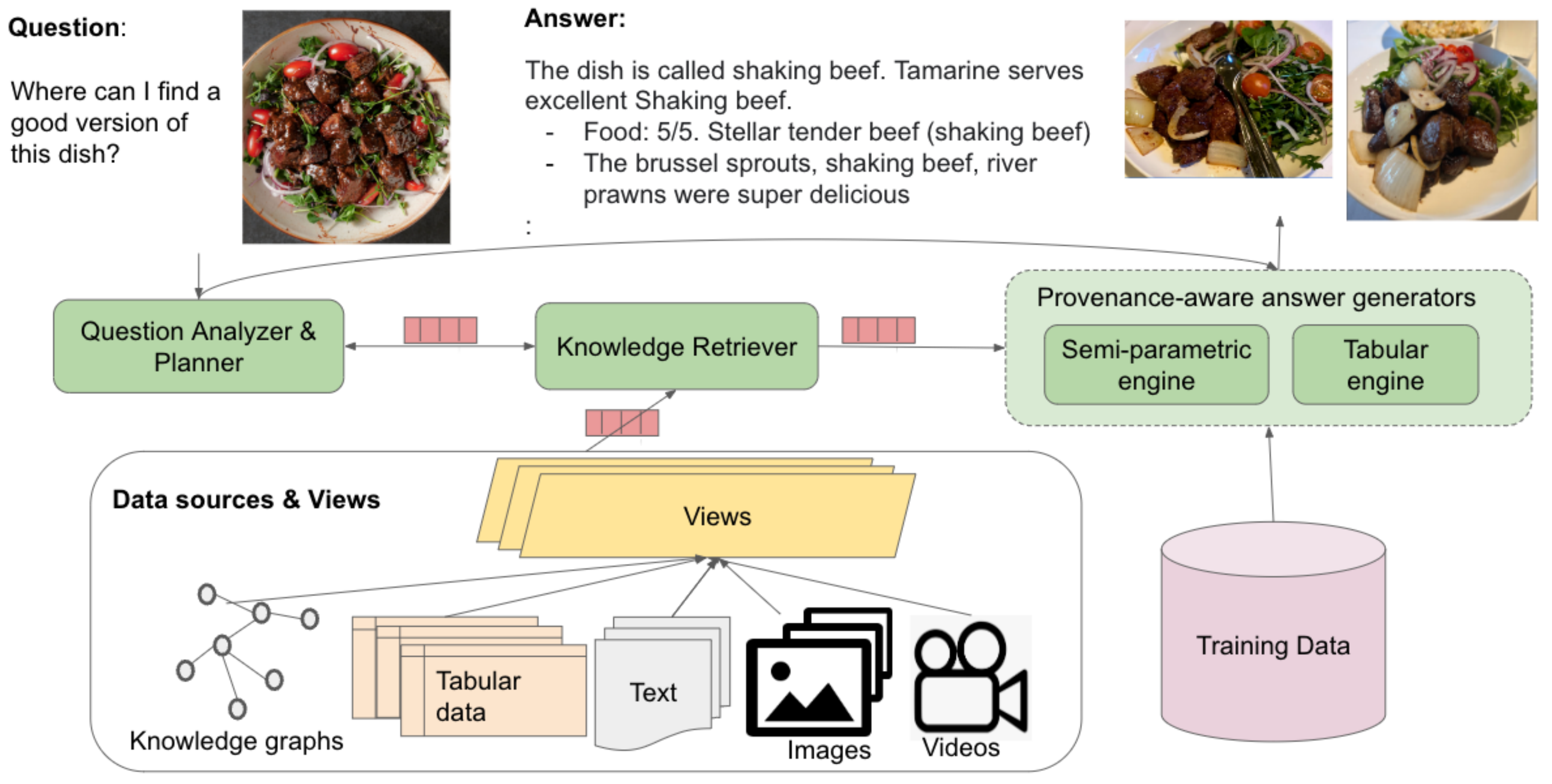}
    \caption{Semi-parametric architectures enhanced with views, a query analyzer \& planner module, and a provenance-aware answer generator. The data sources may be public or private.}
    \label{fig:spa++}
\end{figure*}

\noindent
{\bf PostText architecture~}
\spanew\ enhances the core architecture of semi-parametric models with three 
components: views, a query analyzer \& planner (QAP), and a provenance-aware answer generator (PAG). In addition, all components including the ``traditional'' knowledge retrievers are equipped to manage both structured and unstructured data of different modalities. 

Figure~\ref{fig:spa++} shows the architecture of \spanew{}. Views are synthesized from different types of external data sources (e.g., text, images, videos, and tabular data), which can be public 
or private. When a question is posed in natural language (NL), the QAP module interprets and 
decomposes the question into subquestions whose answers can be composed to obtain an answer to the input question. QAP coordinates with the knowledge retriever to derive the data needed to answer these questions. It also coordinates with the PAG module with its plan so that provenance-aware answers can be returned.

Adding these components raises interesting challenges such as what views should we construct and how do we construct and maintain these views automatically as data sources changes? What is a good plan for deriving an answer and how do we choose among alternative plans? And how do we measure the ``goodness'' of an answer with provenance?

In the remaining sections, we describe the challenges associated with each of these components

\section{Data Sources and Views}

\noindent
{\bf Data Sources~} Most existing work on retrieval augmented language models are focused on text. 
More recently, \citep{chen-etal-2022murag,yasunaga2022ram,sheynin-etal-knndiffusion} has applied \spa{} models on image-text and text-only corpus. The data sources in \spanew{} are multi-modal, unstructured or structured. They can be external public data sources or private ones.  

\smallskip
\noindent
{\bf Views~} 
Views are results computed (not necessarily through SQL queries) from data sources or other views. For example, a view can be a document involving data of different modalities (e.g., an image or a table). Views are powerful constructs for surfacing important and useful associations that are not obvious otherwise, whether they are associations from data within one data source or across multiple data sources. The table in Figure~\ref{fig:example1-multimodalquestion} is a view over 
restaurant reviews from Yelp, Google, and images provided by restaurants. 
This view makes it easier to compute the number of reviews associated with each dish in each restaurant or even across all restaurants. This view also makes it easier to determine the answer as to which dishes has more reviews than Shaking beef at Tamarine. 

Indexes are a special type of views. They associate
an item with its attribute. 
Several implementations of retrieval augmented language models~\citep{guu2020realm,lewis2020rag,izacard2022atlas} already construct indices that associate a document with its nearest neighbors. Recently, GPT-index~\citep{gpt-index-2022} developed a set of APIs for creating data structures that can be traversed using LLMs to answer queries. The data structures are structured indexes and can be used to determine an answer to a question. 

Relational views are extensively used in data warehouses for optimizing queries. 
Indexes and views are typically created by users or database administrators or they can 
be automatically selected~\citep{agrawal-etal-2000automatedview,schnaitter-etal-2007,jindal-etal-2018selectviews} and tuned~\citep{agrawal-etal-2006,brunochaudhuri-2008} to efficiently answer queries of a given workload~\citep{das-etal-2019msazure}, which are queries that are anticipated to be frequently occurring.
In typical settings, a set of views are constructed, usually under a budget constraint such as space, to maximize the queries that can be answered (either directly or through applying a few simple operators on the views) in a given workload. When a new query arrives after the views are constructed, the query optimizer determines the best plan to adopt for computing the answer. 
Queries are directly executed over the views if possible. Otherwise, it falls back to old strategy of answering the query with the data sources. For example, early last year, in anticipation of frequent queries about statistics of past World Cups due to the World Cup 2022 event at the end of the year, a set of views about the different World Cup statistics could have been constructed a priori so that most World Cup related questions can be directly answered using the views. 

We hypothesize that views in \spanew{} can bring similar benefits to question answering. 
The right views will make it easier for the QAP module and the knowledge retriever to discover and obtain relevant data and subsequently for the answer generator to derive the right answers. 
Existing SPAs~\citep{guu2020realm,lewis2020rag,izacard2022atlas} are already
leveraging dense-vector indices to accelerate the retrieval of 
document spans.
In \spanew{} with views being available, it is a natural extension
to annotate each view with a description of its content (e.g., ``{\em Restaurants and highly ranked dishes}''), which would make it even easier for the knowledge retriever to find the relevant data. 
The core challenges in developing views are how do we determine what is a ``right'' set of views to materialize automatically or semi-automatically?
How do we incrementally maintain such views as data sources are updated? These problems are extensively studied in the database community and it will be interesting to explore 
those ideas that transfer to the \spanew. 

The architecture can also be instrumented in such a way that views are the only sources of data for the knowledge retriever (i.e., actual data sources are excluded). Hence, in this case, views act as a gateway that define which parts of the data sources are accessible by the knowledge retriever to answer queries. Finer-grained access control can also be instrumented through views as described in~\citep{bertinosandhu-2005}. 
 With views, it is also possible to enable a finer-grained public-private autoregressive information retrieval privacy system~\citep{arora-etal-2022publicprivate}.




\section{Question Analyzer \& Planner}
\label{sec:qap}

The question analyzer and planner (QAP) module examines the input question and generates a plan, i.e., a sequence of sub-questions whose answers can be combined to form an answer to the input question.
For 
each subquestion in the plan, QAP first checks whether external knowledge is needed. If not, the language model can be used to derive the answer. Otherwise, the subquestion is passed to the knowledge retriever to discover and retrieve relevant data for the subquestion at hand. The results from the knowledge retriever and the plan are passed to PAG (i.e., the rightmost green box in Figure~\ref{fig:spa++}).
It is still an open and challenging question to determine whether a language model can confidently answer a question~\citep{KamathJL20, Si2022RevisitingCF}. Any solution to this problem will help improve the plan generator. 

An example plan from the QAP module for our running example is as follows: (1) find the name of the dish $X$ in the input image, (2) find restaurants that serve $X$, (3) find the top restaurant among the results from (2).  This plan is viable because (a) there is an index associating embeddings of images with the name of the main entity of the image, (b) there exists a view as shown in Figure~\ref{fig:example1-multimodalquestion}, which supports the search for restaurants that serve a particular dish. 
Top answers can be derived by computing the scores of the reviews or approximating it based on the sentiment of the reviews and then ranking the results based on such scores. 
The information from (2) is passed to PAG which will compute the answer along with its provenance. This plan is based on the heuristic to push selection conditions early before joining/combining different data sources if needed.  The conditions in the question are ``good version'' and ``this dish''. In this case, no joins are required as the view already combines the required information in one place. Hence, QAP seeks to first find the name of the dish to narrow down the reviews restricted to this dish. Alternatively, it could also retrieve all good reviews before conditioning on the name of the dish. Yet another plan could be to match the image directly to the images of the view to find the top reviews. Or, it may decide to directly retrieve only top reviews with images similar to the image in the question from the external data sources and condition the answer based on the name of the restaurant mentioned in the reviews.

In all possible plans, the knowledge retriever is responsible for discovering and retrieving the relevant data for the QAP plan. 
In addition to the logic that may be needed for decomposing the question into subquestions, a plan is also needed for composing the subanswers obtained to form an answer to the input question. The plan is shared with the PAG module for deriving the associated provenance.

A fundamental challenge in developing the QAP module is how to derive candidate plans and decide what is the ``best'' plan for answering the question when there are different ways to obtain an answer. Achieving this requires understanding how to compare amongst alternative plans for deriving an answer to the question. This problem bears similarity to query evaluation techniques for database systems (e.g.,~\citep{graefe93}). 
It will be interesting to investigate whether database query planning techniques and ideas can synergize with question understanding and planning techniques (e.g.,~\citep{wolfson2020break, Dunietz2020-ty, zhao-etal-naacl2021, xiong-etal-iclr2021} to develop a comprehensive query planner.
Emerging work such as chain of thought reasoning~\citep{chainofthought-2022}, where a sequence of prompts are engineered to elicit better answers, ReAct~\citep{ReAct2022}, where reasoning and action techniques are applied for deriving an answer, and more recently, work that generates a plan which can call LMs for resolving subquestions~\citep{Cheng-etal-2022binding} are also relevant. These techniques so far are restricted to text and does not compare among different plans. 

Another challenge in the context of NL questions is that while there is a single correct answer to an SQL query over a database, there are potentially many different correct answers to a NL question~\citep{si-etal-2021-whats,min-etal-2020-ambigqa,chen-etal-2020-mocha}. Hence the space of possible plans to derive the ``best'' answer most efficiently is even more challenging in this case.


We are advocating for a system that can reason and compare at least some viable strategies to arrive at a best plan for deriving a good answer efficiently. 
Naturally, one can also train a LM to create a plan. Our belief is that taking a more systematic route to planning can relief the need for the amount of training data required and will also aid provenance generation through its ability to describe the steps it took and the sources of data used in each step to generate an answer.
As we shall explain in Section~\ref{sec:kr}, the cost and accuracy of knowledge retrievers can also play a role in determining what is a better strategy for computing a good answer.

\section{Knowledge Retriever}
\label{sec:kr}

The role of the knowledge retriever is to provide the information that the system lacks in order to fulfill the given task, typically at the inference time.
More importantly, we envision that the knowledge retriever proposed in our framework has the ability to access knowledge stored in different sources and modalities, retrieve and integrate the relevant pieces of information, and present the output in a tabular data view.
The structured output contains raw data items (e.g., text documents, images or videos) and 
and optionally different metadata, such as textual description of each data item.
Such structured output allows downstream (neural) models to consume the retrieved knowledge efficiently and also allows developers and users to validate the provenance conveniently.
Existing information retrieval models mostly focus on a single form of data.
Below we first describe  briefly how knowledge retrieval is done for unstructured and structured data. We then discuss the technical challenges for building a unified knowledge retriever, as well as recent research efforts towards this direction.


\paragraph{Retrievers for unstructured data}

For unstructured data, such as a large collection of documents (i.e., text corpus) or images, knowledge retrieval is often reduced to a simple similarity search problem, where both queries and data in the knowledge source are represented as vectors in the same vector space~\citep{turney2010frequency}. 
Data points that are \emph{close} to the query are considered as \emph{relevant} and thus returned as the knowledge requested.
Traditional information retrieval methods, whether relying on sparse vector representations, such as TFIDF~\citep{Salton1975tfidf} and BM25~\citep{robertson2009probabilistic}, or dense representations, such as LSA~\citep{Deerwester1990IndexingBL}, DSSM~\citep{Huang2013dssm}, DPR~\citep{karpukhin-etal-2020-dense}, are the canonical examples of this paradigm.
Notice that the vector space model is not restricted to text but is also applicable to problems in other modalities, such as image tagging~\citep{weston2011wsabie} and image retrieval~\citep{Gordo2016deep_image_retrieval}. 

\paragraph{Retrievers for structured data}

When the knowledge source is semi-structured (e.g., tables) or structured (e.g., databases), the query can be structured and allows the information need to be defined in a more precise way.
Because the data is typically stored in a highly optimized management system and sometimes only accessible through a set of predefined API calls, the key technical challenge in the knowledge retriever is to formulate the information need into a formal, structured query.
To map natural language questions to structured queries, semantic parsing is the key technical component for building a knowledge retriever for structured data.
Some early works propose mapping the natural language questions to a generic meaning representation, which is later translated to the formal language used by the target knowledge base through ontology matching~\citep{kwiatkowski-etal-2013-scaling,Berant2013SemanticPO}.
Others advocate that the meaning representation should be closely tight to the target formal language~\citep{yih-etal-2015-semantic}, such as SPARQL for triple stores.
Because of the success of deep learning, especially the large pre-trained language models, semantic parsing has mostly been reduced to a sequence generation problem (e.g., Text-to-SQL).
For example, RASAT~\citep{qi-etal2022rasat} and \textsc{Picard}~\citep{scholak-etal-2021-picard}, which are generation models based on T5~\citep{raffel2020t5}, give state-of-the-art results on benchmarks like Spider~\citep{yu-etal-2018-spider} and CoSQL~\citep{yu-etal-2019-cosql}.


\paragraph{Towards a unified knowledge retriever}

As knowledge can exist in different forms, a unified knowledge retriever that can handle both structured and unstructured data in different modalities is more desirable. 
One possible solution for realizing a unified retriever is to leverage multiple single-source knowledge retrievers. 
When a query comes in, the QAP module first decomposes it into several smaller sub-queries, where each sub-query can be answered using one component knowledge retriever.
The results from multiple knowledge retrievers can be integrated and then returned as the final output.
However, several technical difficulties, including how to accurately decompose the question and how to join the retrieved results often hinder the success of this approach.
Alternatively, unifying multiple sources of information in a standard representation, using text as a denominator representation, has been promoted recently~\citep{oguz-etal-2022-unik, zeng-socratic-2022}.
If all data items have a corresponding textual description, it is possible for the knowledge retriever to use only text-based retrieval techniques to find relevant data items once all input entities of non-textual modality have been mapped to their corresponding textual descriptions. 

Such approach circumvents the complexity of managing multiple knowledge stores in different format.
Moreover, with the success of large multi-lingual and multi-modal language models~\citep{Conneau-lample2019cross, Aghajanyan-etal-2022-cm3}, data of different structures or from different modalities can naturally share the same representation space.
While unifying multiple sources of information through representation learning seems to be a promising direction, it should be noted that certain structured information may be lost in the process. 
For example, by flatting a knowledge graph to sequences of (subject, predicate, object) triples, the graph structure is then buried in the textual form.
Whether the information loss limits the retriever's ability to handle certain highly relational queries remains to be seen.

\section{Provenance-aware answer generators}
\label{sec:gen}

\subsection{Semi-Parametric Engine}

Demonstrating the provenance of a QA model prediction should center on identifying the data---whether in training data, retrieval corpora, or input---that is most influential in causing the model to make a particular prediction.
For example, given the question ``{\em who was the first U.S. president?}'', the system should return the correct answer ``{\em George Washington}'' and references to training or retrieval corpora that are---to the model---causally linked to the answer.
If the training or retrieval data included Washington's Wikipedia page, a typical human would expect for this to be included.
However, the requirement we impose is causal and counterfactual: had the model not used that data, the prediction should change.
If the prediction does not change, then from the causal perspective, there may be other data that is either more influential or duplicative (e.g., if \url{whitehouse.gov} is in the training data, it is duplicative).
Next, we describe common semi-parametric models and sketch how this casually-based answer provenance could be obtained and computational challenges to overcome.

Provided an input prompt and retrieved text, semi-parametric models like ATLAS~\citep{izacard2022atlas} or passing documents as prompts to GPT-3~\citep{kasai2022realtime} are adept at generating free-text, short answers.
Likewise, parametric models with flexible input like GPT-3 can be combined with retrievers to achieve a similar goal; alternatively, transformer models can be retrofitted with layers so that passages can be integrated in embedding space~\citep{borgeaud2021retro}.
While retrieval-augmentation is no catch-all panacea to model hallucination, it does mitigate the problem~\citep{shuster2021hallucinate}.
Additionally, models' explanations can make it easier to know when to trust models and when not to~\citep{feng2022explain}.

In the case of QA models that take question plus retrieved text as input, there are several options.
First, the model could provide several alternative answers which provide insight into the distribution of model outputs, rather than just a point estimate.
Second, the model could provide a combination of feature-based explanations such as token saliency maps and the model's confidence in a correct answer~\citep{wallace2018trick}.
When combined, they can jointly influence the degree to which humans trust the model~\citep{lai2019explain}.
However, to provide a complete account of model behavior, we must return to the training of model and the data used.
In short, we endeavor to identify the combination of input, training data, and retrieved text that caused the model to produce the distribution of outputs (i.e., answer(s)).
This is, of course, challenging due to scale of language model training data like C4~\citep{raffel2020t5} and the Pile~\citep{gao2020pile} and that establishing causal---and therefore more faithful---explanations of model behavior is difficult.
Training data attribution is one promising idea in this direction---it uses gradient and embedding based methods to attribute inference behavior to training data~\citep{akyurek2022tracing}.
For example, influence functions~\citep{hampel1974influence,han2020influence} and TracIn~\citep{pruthi2020tracin} link predictions to specific training examples, but are computationally expensive and are approximate rather than exact solutions.
To firmly establish a causal connection, one could fully re-train the model without the identified training examples, but this is prohibitively expensive in practice.
Future development of efficient training data attribution, combined with methods like interpretations of input plus retrieved data, is a promising direction towards more complete explanations of model predictions.

\subsection{Tabular Engine}


As described at the end of Section~\ref{sec:qap}, the knowledge retriever will pass on the data obtained to PAG. The QAP module will pass information about its plan to PAG. If the data obtained is tabular and a SQL query is generated, the information is passed to the tabular engine of PAG to compute the required answer(s). 
The recent advances in Text-to-SQL~\citep{DBLP:conf/acl/WangSLPR20,DBLP:conf/acl/Zhao0PP22} provide 
a good technical foundation for generating such SQL queries.

In most cases, it is not difficult to understand the correspondence between the natural language question and the SQL query that is generated. Once the SQL query is obtained, provenance can be systematically derived. In databases, the notion of provenance is well-studied~\citep{whyhowwhere2009} for a large class of SQL queries; from explaining why a tuple is in the output (i.e., the set of tuples in the database that led to the answer), where a value in a tuple is copied from (i.e., which cell in the source table is the value copied from)~\citep{whywhere01} to how that tuple was derived, which is formalized as semirings~\citep{semiring2007}, a polynomial that essentially describes conjunction/disjunction of records required materialize a record in the result. Database provenance has also been extended to aggregate queries~\citep{provenanceaggregates2011}. Since one can derive the mapping between the input question and the SQL query that is generated and also derive the provenance from the data sources based on the SQL query, it becomes possible to understand how the input question led to the answers given by \spanew{}.

Putting all together, \spanew{} first explains that the name of the image (i.e., ``{\em a good version of this dish}'') referred in question is Shaking beef. It then shows the SQL query that is generated for the question ``{\em Where can I find a good version of Shaking beef}'' and the ranking function used for ranking the rows of restaurants with reviews for the dish Shaking beef. For our running example, the answer is obtained from the first row of the table in Figure~\ref{fig:example1-multimodalquestion}. Specifically, the answer is summarized from the column {\em Dish} and {\em Review snippets/embeddings}. The actual snippets are found following the provenance links captured in the column {\em Provenance}. A more direct relationship between the summary and the actual review snippets can also be established~\citep{carmeli-etal-2021}. 

The success of this approach depends on how far we can push database provenance systematically as SQL queries can still be far more complex than what is investigated in past research (e.g., complex arithmetic and aggregate functions involving also negation, group filters, and functions over values of different modalities).
As an alternative to executing the SQL query over the tables obtained, the tabular engine can also choose to deploy table question answering (tableQA) methods where a model directly searches the tabular data for answers based on the input question~\citep{sun2016tableQA}. 
Tapas~\citep{DBLP:conf/acl/HerzigNMPE20} and Tapex~\citep{DBLP:conf/iclr/LiuCGZLCL22}
are two example solutions for tableQA that formulates tableQA as sequence 
understanding/generation tasks.
Like other recent tableQA works~\citep{DBLP:conf/naacl/GlassCGCKCSPBF21,
DBLP:conf/naacl/HerzigMKE21},
they consider the problem of computing the answer from a single input.
It will be interesting to explore how to explain the results obtained using tableQA methods and how tableQA methods can be extended to handle multi-hop questions where the answer may span multiple tables or involve different types of aggregations, reasoning and modalities.



\section{Preliminary Findings}
\label{sec:expfindings}
To test our hypothesis that views are valuable for answering queries, especially queries that involve counting or aggregation, we have implemented a first version of \spanew{}\footnote{PostText source code will be made available soon.} and compared it against some QA baselines. 

The current implementation of \spanew{} assumes views over the underlying data are available in tabular format. The QAP module simply routes the query to a view-based engine (VBE) or a retrieval-based engine (RBE) to answer the query. VBE picks the best view and translates the natural language query into an SQLite query against the view using OpenAI's {\tt gpt-3.5-turbo/gpt-4} model. It then executes the SQLite query against the view to obtain a table result which is then translated into English as the final answer.
VBE also analyzes the SQLite query to compute the provenance of the answers. At present, it does so by simply retrieving all tuples that contributed to every (nested) aggregated query that is a simple (select-from-where-groupby-having clause) and does not handle negations. An example of the VBE process is described in Appendix~\ref{sec:vba}. RBE is implemented with Langchain's {\tt RetrievalQAwithSources} library. It first retrieves top-$k$ documents that are relevant for the query and then conditions its answer based on the retrieval. The answer and the ids of the retrieved documents are returned.

For our experiments, we use the 42 multihop queries over 3 synthetic personal timelines of different sizes from TimelineQA's benchmark~\cite{timelineqa}. The personal timelines model the daily activities (e.g., the trips made, things bought, people talked to) of a person over a period of time.  We create a view around each type of activity (e.g., trips, shopping, daily\_chats) for VBE. For further comparison, we also ran Langchain's SQLDatabaseChain (DBChain) to perform QA over the same VBE views. Furthermore, we ran it over timelines loosely structured as a binary relation of (date,description) pairs (called DBChain (no views)). 
We compared the returned answers against the ground truth answers by grading them on a scale of 1-5, with a LLM, where 5 means the returned answer has the same meaning as the ground truth answer (the grading scheme is described in the Appendix~\ref{sec:gradingscheme}).

\begin{table}[!t]
\small
\begin{tabular}{|c|c|c|c|>{\centering\arraybackslash}p{2.7cm}|}
\hline
  & {\bf VBE} & {\bf RBE} & {\bf DBChain} & {\bf DBChain (no views)} \\
\hline
\hline
 S & 3.45 & 2.81 & 3.37 & 2.72 \\
 M & 3.79 & 2.69 & 3.28 & 2.61 \\
 L & 3.11 & 2.44 & 2.95 & 1.95 \\
 \hline
\end{tabular}
\caption{Results with GPT-3.5-turbo. Sizes of (S)mall, (M)edium, (L)arge are 1.1MB, 2.4MB, and 5.6MB respectively.}
\label{tbl:gpt3.5}
\end{table}

\begin{table}[!t]
\small
\begin{tabular}{|c|c|c|c|>{\centering\arraybackslash}p{2.7cm}|}
\hline
  & {\bf VBE} & {\bf RBE} & {\bf DBChain} & {\bf DBChain (no views)} \\
\hline
\hline
 S & 3.33\textsuperscript{*} & 2.10\textsuperscript{*} & 2.14\textsuperscript{*} & 1.10\textsuperscript{*} \\
 M & 3.55 & 1.93 & 2.35 & 1.51\textsuperscript{*} \\
 L & 3.08 & 2 & 1.97 & 1.11\textsuperscript{*} \\
 \hline
\end{tabular}
\caption{Results with GPT-4. \textsuperscript{*} indicates that timeouts or API errors were encountered during experimentation.}
\label{tbl:gpt4}
\end{table}

Our results are shown in Tables~\ref{tbl:gpt3.5} and \ref{tbl:gpt4}. 
Across both tables, the results on DBChain vs. DBChain(no views) reveal that adding some structure (in this case adding views) is crucial for better performance.  
Although the benchmark is a relatively small dataset, the scale of the timelines already reveals an impact on the accuracy across all QA systems. For DBChain, the drop in accuracy as the size increases because it sometimes relies on generating SQL queries that return all relevant records and passing all the records to the language model to compute the aggregate. When the results returned are large, which tends to be the case for larger timelines, the token limit of the LLM is often exceeded. 
VBE has a similar downward trend. It tends to generate queries that push the aggregates to the SQL engine and hence, avoids the issue of exceeding the token limit of the language models for many cases encountered in DBChain. Still, as the timeline gets larger, the result returned by the generated SQL query tends to be bigger and when these results are passed to the verbalization component to compose an answer in English, this may sometimes exceed the token limit of the language model. We also found that on a handful of cases, it so happens that the SQL query generated for L is invalid compared with those generated for the sparse dataset.

The scores of RBE is relatively stable across all data densities. But overall, it tends to score lower compared with VBE and DBChain . This is because RBE relies on retrieving the top $k$ documents from an index to condition the answers upon, regardless of the size of the timeline. However, these retrieved documents may not contain all the necessary information for answering the question in general. Even though the grading scores may not reveal this, the answers tend to be ``more wrong'' for aggregate queries over a larger timeline.

\section{Conclusion}
\spanew\ enhances the core ideas of semi-parametric architectures with views, a query analyzer \& planner, and a provenance-aware answer generator. Our initial results indicate that \spanew{} is more effective on queries involving counting/aggregation when we provide structured views to facilitate computation.  
We plan to further develop and investigate \spanew{} to automatically determine what views to construct, how does one generate plans and compare amongst plans, and how can one measure the quality of answers with provenance.


\section*{Limitations and Ethical Considerations}
We point out the limitations of large language models (costly to train, deploy, maintain, hallucinate, opaque). The vision of \spanew\ shows promise of less costly training, maintenance, and more explainability. However, no actual system is built yet to validate these claims and it is also not clear that a system with \spanew\ architecture will be easier to deploy since it has more components. 

\bibliography{bib/journal-full,bib/pedro,bib/bib,bib/anthology}
\bibliographystyle{acl_natbib}

\appendix

\section{Appendix}
\label{sec:appendix}

\section{View-based QA}
\label{sec:vba}
Example run of \spanew{} with the query "{\em When was the last time I chatted with Avery?}":

This query is first matched against a set of available views and the best one is picked if there is sufficient confidence. In this case, the view {\tt daily\_chat\_log} is selected.

The query is first translated into an SQLite query:

\small
\begin{center}
\begin{verbatim}
SELECT MAX(date)
FROM daily_chat_log
WHERE friends LIKE '%Avery%'
\end{verbatim}
\end{center}
\normalsize

The SQLite query is then cleaned and ``relaxed''. For example, on occasions, an attribute that does not exist is used in the query even though this happens rarely. In this case, no cleaning is required. The conditions over TEXT types are also relaxed. We convert equality conditions (e.g., {\tt friends = 'Avery'}) to {\tt LIKE} conditions (e.g., {\tt friends LIKE '\%Avery\%'}) and further relax LIKE condition with a user-defined {\tt CLOSE\_ENOUGH} predicate. 

\small
\begin{verbatim}
SELECT MAX(date)
FROM daily_chat_log
WHERE (friends LIKE '%Avery%' OR 
       CLOSE_ENOUGH('%Avery%', friends))
\end{verbatim}
\normalsize

The above query is executed and the results obtained is shown below.
We then verbalized an answer based on the table result.
\smallskip
\noindent
{\bf Result:}
[('2022/12/26')]

\medskip
\noindent
{\bf Returned answer (verbalized)}: 
{\em The last time I chatted with Avery was on December 26, 2022.}

We observe that Langchain's SQLDatabaseChain provides a very similar functionality of matching an incoming query against available tables and generating an SQL query over the matched tables. However, SQLDatabaseChain does not clean or relax query predicates, and requires one to specify a limit on the number of records returned. Furthermore, it does not compute the provenance of the answer obtained, as we will describe in the next section. As we also described in Section~\ref{sec:expfindings}, view-based QA generally outperforms SQLDatabaseChain because of its ability to push aggregates to the database engine instead of relying on the language model to aggregate the results (after using the database engine to compute the relevant records for answering the query.

\medskip
\noindent
{\bf Provenance queries}: PostText generates queries to retrieve records that contributed to the answer returned above. It does so by analyzing every {\tt select-from-where-groupby-having} subquery in the generated query to find tuples that contributed to every such subquery. For example, the following SQL queries are generated to compute provenance. 

\small
\begin{verbatim}
SELECT name 
FROM pragma_table_info('daily_chat_log') 
where pk;

q0: 
SELECT eid 
FROM daily_chat_log
WHERE (friends LIKE '%Avery%' OR 
       CLOSE_ENOUGH('%Avery%', friends))
\end{verbatim}
\normalsize

The first query above returns the key of the table and the second retrieves the keys from the table that contributed to the returned answer. 

\small
\begin{verbatim}
[('q0', ('e152',)), ('q0', ('e154',)), ('q0', 
  ('e169',)), ('q0', ('e176',)), ...]
\end{verbatim}
\normalsize

\section{Grading scheme}
\label{sec:gradingscheme}
The following is our grading scheme used for grading the answers generated by different systems against the ground truth answer:

\begin{itemize}
\item 5 means the systems's answer has the same meaning as the TRUE answer.
\item 4 means the TRUE answer can be determined from the system's answer.
\item 3 means there is some overlap in the system's answer and the TRUE answer.
\item means there is little overlap in the system's answer and the TRUE answer.
\item 1 means the system's answer is wrong, it has no relationship with the TRUE answer.
\end{itemize}

\end{document}